\documentclass{article}
\usepackage{acra}
\usepackage{graphicx}
\usepackage{multirow}
\usepackage{adjustbox}
\usepackage{amsmath}
\usepackage{url}
\usepackage{hyperref}
\usepackage{array}

\title{Does ChatGPT and Whisper Make Humanoid Robots More Relatable?}
\author{Xiaohui Chen$^{1*}$, Katherine Luo$^{1}$, Trevor Gee$^{1}$,  Mahla Nejati$^{1**}$ \\ Centre for Automation and Robotic Engineering Science, The University of Auckland, New Zealand$^{1}$.\\
cxia813@aucklanduni.ac.nz$^{*}$, m.nejati@auckland.ac.nz$^{**}$}

\begin{document}

\maketitle

\begin{abstract}


Humanoid robots are designed to be relatable to humans for applications such as customer support and helpdesk services. However, many such systems, including Softbank’s Pepper, fall short because they fail to communicate effectively with humans. The advent of Large Language Models (LLMs) shows the potential to solve the communication barrier for humanoid robotics. This paper outlines the comparison of different Automatic Speech Recognition (ASR) APIs, the integration of Whisper ASR and ChatGPT with the Pepper robot and the evaluation of the system (Pepper-GPT) tested by 15 human users. The comparison result shows that, compared to the Google ASR and Google Cloud ASR, the Whisper ASR performed best as its average Word Error Rate (1.716\%) and processing time (2.639 s) are both the lowest. The participants' usability investigations show that 60\% of the participants thought the performance of the Pepper-GPT was "excellent", while the rest rated this system as "good" in the subsequent experiments. It is proved that while some problems still need to be overcome, such as the robot's multilingual ability and facial tracking capacity, users generally responded positively to the system, feeling like talking to an actual human.
\end{abstract}

\section{Introduction}

As technology rapidly advances, the need to streamline the interaction between humans and machines is becoming more critical. This problem has become so important that an entire sub-field within robotics called Human-Robot interaction (HRI) has emerged to address it. The inability of humans to interact with machines effectively is a known technology bottleneck \cite{sharma1998toward} that prevents us from getting the full benefit from these systems.

Hardson and Pyla et al. defined user experience (UX) as the impact of the machine on the user, including simplicity, intuitiveness, usefulness, and frustration levels during and after the interaction \cite{Hartson2012}. Ensuring a favourable user experience during interactions between humans and robots is critical in ensuring that these machines bring actual value to our lives \cite{Khan2018,Alenljung2017}.

The well-known humanoid social robot Pepper, created by SoftBank Robotics, is celebrated for its extensive interactive features and versatile functions, such as speech recognition, pre-coded dialogue, gesture, facial tracking capabilities, etc. However, if the robot is envisioned to become more adaptable and human-like in communication, its current abilities might fall short of expectations \cite{Gardecki2018}. 

Existing studies have found that despite developers adding essential features to the Pepper robot, it still faces challenges when performing tasks. Noticeable delays and errors in Pepper's language processing appear during responses, which cause a significant impact on participants and make users feel challenged to stay engaged during subsequent session phases \cite{Matulik2020}. 

The Pepper robot does come with a tool for managing communication. This dialogue-based tool allows developers to embed scripted conversations into Pepper. The main disadvantage of this approach is that interactions are pretty limited and are often unnatural to users who want deeper elaborations or seek clarification on specific points. Poorly designed scripts lead Pepper to misinterpret queries or fail to provide adequate answers to questions \cite{Foster2019}. 

In addition, users reported difficulties in the robot's ability to understand their speech, often requiring multiple attempts for the robot to comprehend their input \cite{Corrales-Paredes2023}. The Pepper robot's built-in speech recognition API can only recognize predefined specific phrases, so its ability to understand natural speech is inadequate.

The recent growth in sophisticated Large Language Models (LLMs) such as Chat GPT is utilised to address these problems. These systems provide articulate, detailed responses according to arbitrary (unscripted) questions \cite{Wu2023}.

Given its exceptional performance, it is postulated that users should be able to engage in more genuine and contextually fitting conversations with ChatGPT than from the current Pepper robot's dialogue tool. Furthermore, enhancing the speech recognition capabilities of the Pepper robot is crucial to ensure an accurate and complete transcription of audio input. This improvement can minimise the probability for the robot to misunderstand user speech, ultimately enhancing the overall performance of the robot and promoting the UX.

This paper outlines the methodology of integrating the Pepper robot with the Whisper and GPT API (the system is called Pepper-GPT) and the results from subsequent experiments of human participants and Pepper-GPT interactions.

\section{Related Work}

In studies examining user experiences with humanoid robots, the predominant approach has been to conduct human trials, involving participants in specific real-life scenarios where they interact with these robotic entities. For instance, there have been instances of robots designed for healthcare receptionists \cite{Johanson2020}, which offer valuable opportunities to gain firsthand insights into the user experience. Typically, researchers employ questionnaires to gather participants' feedback on their feelings, perceptions, and attitudes toward the robots, with these questionnaires thoughtfully designed to evaluate various aspects of the interaction, including user satisfaction, comfort, trust, and the perceived utility of the robot \cite{Corrales-Paredes2023}.

Moreover, to enhance the communication capabilities of robots like Pepper, extensive research has focused on enhancing its speech recognition and natural language processing abilities. For example, in a study that utilized Pepper Robot for front desk applications, researchers integrated Google Cloud Speech Recognition to enhance the robot's native speech recognition \cite{Gardecki2018}. While they did not specify the exact NLP methods used, they proved its adequacy for reception tasks. Their quantitative analysis on Google Cloud showed satisfactory results, with an average response time of 1.031 seconds in noisy conditions and 0.887 seconds in quiet conditions at a microphone distance of 0.4m \cite{Gardecki2018}. However, it is worth noting that they did not employ the Word Error Rate (WER) to measure the accuracy of Automatic Speech Recognition (ASR) \cite{Vallath2022}, leaving room for further investigation into the ASR's actual accuracy.

The following sections delve into the examination of the existing literature concerning speech recognition and techniques used in natural language processing.

\subsection{Speech Recognition}
After reviewing various comparisons of ASR systems' Word Error Rates (WER), three ASR models emerged as standout performers. Google, a supported Google ASR in Python library, showed the smallest average WER at 20.63\% among IBM, Google and Wit \cite{Filippidou2020}. Google Cloud, a powerful and versatile cloud-based service offered by Google Cloud, had an average WER at 12.16\% \cite{PiresMagalhaes2022} and Whisper, a new ASR system developed by OpenAI in 2023 \cite{openAIDoc}, which it outperformed all its competitors with its WER ranging from an impressive low of 3.523\% to a maximum of 19.6\% \cite{radford2023robust}. Notably, while the study highlighted Whisper's exceptional performance, it did not specify all the competitors it outperformed.

However, there is a scarcity of direct quantitative comparisons between these three ASR systems and a limited variety of accent tests among them. This is primarily because most evaluations have been conducted with speakers from the same language country. Therefore, a subsequent evaluation was conducted to determine the superior performer among Whisper, Google, and Google Cloud ASR systems. The outcome highlighted Whisper as the leader in terms of both accuracy and response time.

\subsection{Natural Language Processing}
Similarly, another study showcased an approach closely aligned with our proposed method \cite{Billing2023}. In this study, a research team successfully integrated Pepper with OpenAI's large language models, specifically GPT-3. They replaced Pepper's native speech recognition with Google Cloud Speech-to-Text technology, enabling Pepper to engage in open verbal dialogues with users, thus expanding its conversational capabilities beyond predefined commands. 

However, now, with the adoption of much up-to-date advanced language models such as GPT-3.5 and GPT-4.0, there is more potential to elevate conversational abilities further \cite{Ditto2023}. The gpt-3.5-turbo model is particularly advantageous due to its capacity to handle multi-turn conversations, facilitating the processing of a sequence of messages as input. This marks a significant improvement over the GPT-3 model, which was limited to single-turn text prompts \cite{Brown2023}.

Additionally, gpt-3.5-turbo maintains competitive performance compared to GPT-4 and balances efficiency with budget-friendliness, making it a suitable choice for projects prioritizing cost efficiency and lower latency \cite{openAIDoc}.

In conclusion, ongoing research and technological advancements in ASR and NLP have paved the way for Pepper to become an even more effective and engaging communication tool, offering valuable interactions and assistance to many users. The future holds exciting possibilities for further improving the robot's capabilities, making it an even more indispensable asset in various applications.

\section{Methodology}
\begin{figure}[!htb]
    \centering
    \includegraphics[scale=0.225]{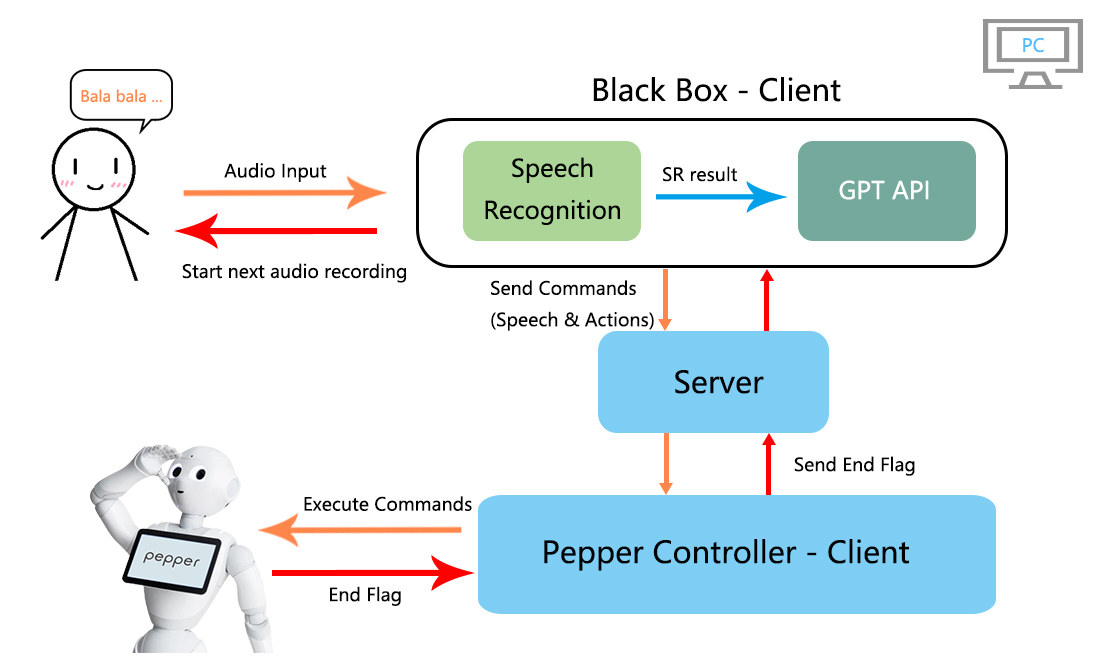}
    \caption{The Workflow of Pepper-GPT.}
    \label{fig:workflow}
\end{figure}

This section outlines the methodology employed in the design and implementation of Pepper-GPT. The design architecture consists of two core programs: BlackBox (Speech Recognition System and the GPT model) and PepperController (Fig.~\ref{fig:workflow}). In addition, a client-server model with TCP/IP protocol is used for data transmissions between these two programs. 

BlackBox incorporates the Whisper Automatic Speech Recognition(ASR) system and the gpt-3.5-turbo language model. The PepperController is a command centre designed for executing physical action commands. The BlackBox utilizes the Whisper ASR and gpt-3.5-turbo capabilities to facilitate responsive and contextual-relevant communication in this framework. Meanwhile, the PepperController directs the robot's actions. 

The design of Pepper-GPT overcomes the limitations of artificial intelligence models that cannot engage in physical activities due to their lack of physical presence.

\subsection{BlackBox}
BlackBox can be divided into two main modules: the speech recognition module and the GPT module. Through these two modules, BlackBox records the user's audio input, recognizes the speech content, and generates accurate action commands or contextual replies via the GPT-3.5 model. The generated result is then sent to PepperController for execution.

\subsubsection{Speech Recognition Module}
To enhance the robot's speech recognition to transcribe the voice input content of users fully, three different ASR models were tested, and the Whisper ASR system was selected because of the best performance. The Whisper ASR system developed by OpenAI was famous for its robustness and exceptional performance \cite{radford2023robust}. Its outstanding performance makes it a convincing choice to improve the Pepper robot's speech recognition ability compared to its native API. This choice aims to significantly enhance the robot's ability to understand speech, facilitating more effective user interactions.

Among the array of available Whisper models, the Whisper Small model prominently stands out for its exceptional performance in maintaining an outstanding equilibrium between processing time, resource utilization, and transcription accuracy. Collectively analysing the data of model sizes, Word Error Rates (WER), and time consumption, it is evidenced that even though the Whisper Small model maintains a moderate level of performance, its comprehensive performance is outperforming both smaller and larger sizes models \cite{radford2023robust}. This collective assessment solidifies its position as the top performer among all models for English speech transcription, demonstrating its exceptional speech recognition capabilities.

The design of the speech recognition module incorporates a microphone that activates when it detects human sounds, initiates the recording process, and cleverly ceases when there is silence. It is essential to prevent empty audio while making smooth conversations as this kind of audio file would be transcribed into “Thank you.” or “Thank you for your watching.” by Whisper, leading to the transmission of the wrong message into the GPT module for generating responses. Therefore, the Silero VAD model \cite{SileroVAD} is also integrated for detecting human voices to address this problem. 

The audio recordings are saved as files and then transcribed into text by the Whisper Small model, effectively converting spoken words into written form. When the Whisper Small model successfully transcribes the audio input, the resulting text is forwarded to another critical component, the GPT module. The GPT module takes on the role of comprehensive content analysis and response generation.

However, there may be instances where the Whisper Small model encounters difficulties and returns an empty transcription, indicating a transcription failure. In such cases, the system is designed to respond proactively. It sends a command to the Pepper robot, which makes the robot give users a gentle prompt, encouraging them to speak again. This ensures that user interactions remain smooth and frustration-free even in the face of occasional technical problems.

\subsubsection{GPT Module}
The gpt-3.5-turbo model is central to the design architecture and serves as the engine responsible for generating responses. The reason for selecting the gpt-3.5-turbo model is its remarkable comprehension and text generation capabilities, which allow it to produce natural human-like style responses, aligning perfectly with the purpose of creating highly freedom reply content in the Pepper-GPT design. The model's ability to understand user inputs and generate relevant, genuine, and engaging conversations emphasizes its significance in advancing the overall user experience \cite{openAIDoc}.

\begin{figure}[!htb]
    \centering
    \includegraphics[scale=0.4]{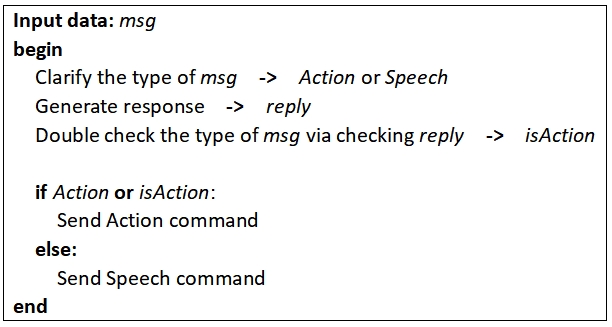}
    \caption{Pseudo code for GPT module.}
    \label{fig:pesudo-code}
\end{figure}

Fig.~\ref{fig:pesudo-code} summarizes the command generation process in the GPT module by pseudo-code. Upon receiving the transcribed text from users via the speech recognition module, the GPT module will first analyze the dialogue content from the user and enter different processing modes according to the analysis results, termed the action mode and the speech mode, respectively.

In action mode, the primary objective of the GPT module is to convert the user’s content into corresponding action commands for the Pepper robot to execute. The GPT-3.5 model does not receive the content directly to generate responses. Instead, it is asked to identify action-related keywords within the dialogue content and transform them into precise physical action commands. 

Conversely, in speech mode, the user’s input is aimed to initiate or continue the conversation with the Pepper robot. In this mode, the GPT-3.5 model reverts to its role as an interlocutor. It receives the dialogue content and generates contextual-relevant replies, ensuring engaging and meaningful user interactions. 

It is worth noting that the analysis from the GPT-3.5 model may occasionally result in misinterpretations of user content, cognitive the physical action command as speech command and vice versa, leading the GPT module into a wrong mode. Therefore, there is a double-check function designed. This function sends the original users' input to the GPT-3.5 model and checks its reply by searching the keywords. It intervenes and makes corrections if the generated responses demonstrate that the model makes a wrong distinction. 

The complex process ensures that users' inputs are dealt with precisely, whether they aim to order the Pepper robot to do physical actions or engage in conversational dialogue. The design of the GPT module prioritizes user intent and the ability of the robot to respond appropriately, enhancing the overall interaction between the user and the robot.

\subsection{PepperController}
The primary purpose of the PepperController is to serve as the command centre for the Pepper robot, executing various physical actions or making conversational replies based on the commands and data received from the BlackBox. The PepperController connects the digital world of command data and the tangible world of the robot, ensuring that the Pepper robot can respond appropriately to user interactions. 

Both actions and speech commands are executed by the Naoqi ALAnimatedSpeech agent. For speech commands, the PepperController can convert text received from the BlackBox into speech. Also, the Pepper robot is set as the contextual mode in its initialization state via the ALSpeakingMovment agent, enabling it to do specific animations autonomously when detecting keywords in the content of replies. This setting makes the Pepper robot more dynamic and attractive. 

Besides, there is a dataset designed for action commands where all actions the Pepper robot can execute are pre-coded and stored. The PepperController finds the corresponding operations based on the received physical action commands for execution. In addition, since the time used to transcribe speech input and generate responses from the GPT model cannot be ignored, it is necessary to command the Pepper robot to do some transition animation, like making some thinking behaviours during the dealing time to make the interaction process more fluent.

\subsection{Data Transmission}
The Naoqi Python SDK provides developers with an interface to build Python modules that have the capability to operate remotely or on the robot \cite{Python_SDK}. It is compatible exclusively with Python 2.7, while both the GPT-3.5 and Whisper APIs demand work with Python 3. Consequently, the BlackBox and PepperController must operate using distinct Python versions, emphasizing the essential of data transmission between these two programs. 

The design architecture of Pepper-GPT adopts a client-server model for data transfer. The BlackBox and the PepperController serve as clients in this model, taking responsibility for different functionality implementations. After dealing with user inputs, the BlackBox client sends the message to the server, including action commands or speech content. The server subsequently forwards the message to the PepperController client. After the Pepper robot completes the commands, an end flag is sent back from the PepperController client to the server, signalling the BlackBox to prepare for the next audio input recording. Different prefix symbols are used to help the server identify the purpose and receiver of the messages.

To ensure stable and reliable data transfer between the BlackBox and the PepperController, the Pepper-GPT leverages the TCP/IP protocol in data transmission. If the sender does not receive a confirmed signal in a reasonable round-trip time, the TCP protocol automatically resends the data to prevent potential data loss. Hence, compared to the UDP protocol, TCP minimizes data loss and guarantees the consistent exchange of messages between clients, making it an optimum solution for data transmission.

\section{Evaluation}
In this section, two comparative analyses were conducted; one was between the selected speech recognition API and others, and the other was the final results of our experiment, which provided an overview of the design of our final experiment and offered an in-depth examination of the quantitative results obtained from our experiment. 

\subsection{Speech Recognition Evaluation
}
Before the experiment, two tests were conducted among three speech-to-text APIs to select the fastest and most accurate speech recognition model. 

\subsubsection{Measurement Metric}
Word Error Rate (WER) was selected as the performance metric because it is the most widely accepted measure for evaluating the accuracy of such systems \cite{Egas-Lopez2020}. It is defined as in equation ~\ref{equation:1}, where S = the total number of Substitution errors refers to when one word is replaced by another, D= the total number of deletion errors refers to where words are omitted from the input, I = the total number of Insertion errors refers to the unintentional addition of words, and $N_1$ = total number of reference words \cite{Filippidou2020}. 
\begin{equation}
        {WER = \frac {S \ + \ D \ + \ I} {N_1}}
        \label{equation:1}
\end{equation}
Recognition time is another metric used to evaluate their performance. Recognition time measures the speed at which these models transcribe spoken language into text. This metric is crucial in real-world applications where timely and efficient speech-to-text conversion is essential.
\subsubsection{Dataset}

The 'Speech Accent Archive' dataset on Kaggle \cite{TATMAN2017} was selected specifically because it comprises speakers from 177 countries, each of whom uttered the same sample sentence in English. This dataset's global representation ensures a diverse range of English accents, making it an ideal choice to assess the performance and adaptability of the chosen speech recognition models. The selected accents included the five native English-speaking countries[United States, United Kingdom, Australia, Canada, and New Zealand] and seven non-native English-speaking countries/regions[India, China, Philippines, Africa, Arabic, France, and Spain] in the initial test. This selection allows for a comprehensive assessment of the speech recognition models' adaptability to diverse accents and linguistic backgrounds, ensuring their real-world applicability in global communication scenarios.

In the second test, the 'daily-dialog' dataset, readily available on Hugging Face \cite{LiYanranandSuHuiandShenXiaoyuandLiWenjieandCaoZiqiangandNiu2017}, was used. This test aimed to evaluate the speech recognition models' proficiency in recognizing and transcribing everyday conversational dialogues. Five different conversation scenarios were selected to ensure that the chosen speech recognition models excel in practical, real-world applications where common dialogues are encountered.

\subsubsection{Results}
\begin{figure}[!ht]
    \centering
    \includegraphics[scale=0.47]{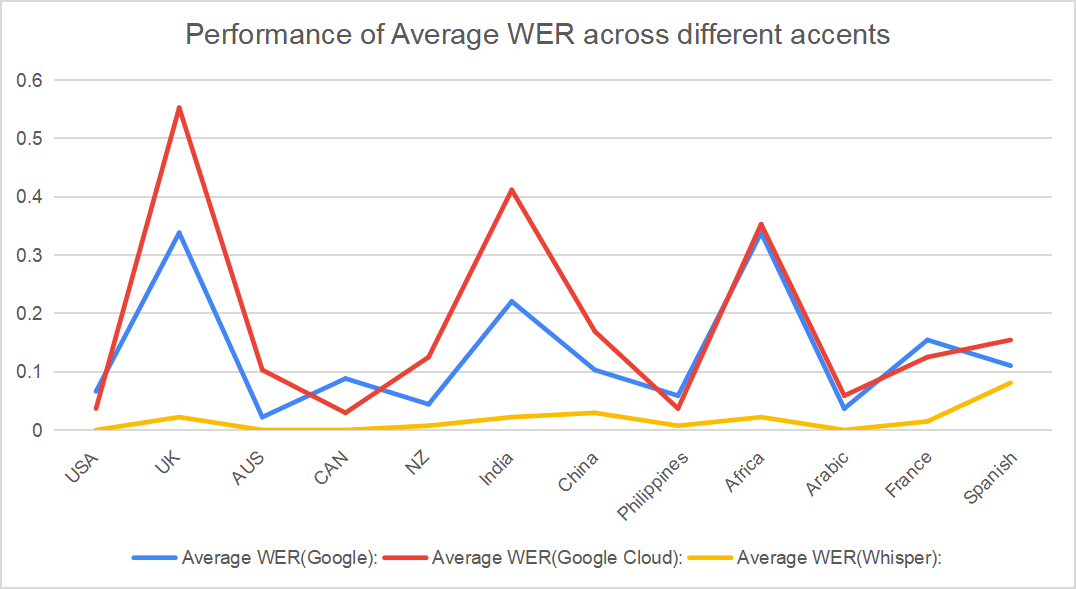}
    \caption{Performance of Average WER across different accents.
}
    \label{fig:4}
\end{figure}

In our initial English accent test, each country was divided into two groups, resulting in 24 data groups. For each group representing a specific country, their speech data were processed using the three speech-to-text APIs and were assessed in terms of both average word error rate (WER) and average recognition time. The results of the three speech-to-text APIs' performance concerning average WER are depicted in Fig.~\ref{fig:4}. A lower WER signifies higher accuracy. Consequently, Whisper achieved the lowest WER among the three, consistently yielding results close to zero, indicating that it correctly transcribed all words. 

The USA consistently achieved the lowest Word Error Rate (WER) among English-speaking countries across all three speech recognition APIs. It is worth noting that native UK English speakers had the highest WER compared to Google and Google Cloud across all accents, with average rates of 0.338 and 0.553, respectively. As native English-speaking countries were expected to have a higher recognition rate than non-native countries.

Among non-English-speaking countries, the Indian accent presented the most challenging understanding, with an overall average WER rate of 0.218. Conversely, the Arabic and Filipino accents were the easiest to understand, boasting an overall average WER rate of just 0.032.

\begin{figure}[!ht]
    \centering
    \includegraphics[scale=0.40]{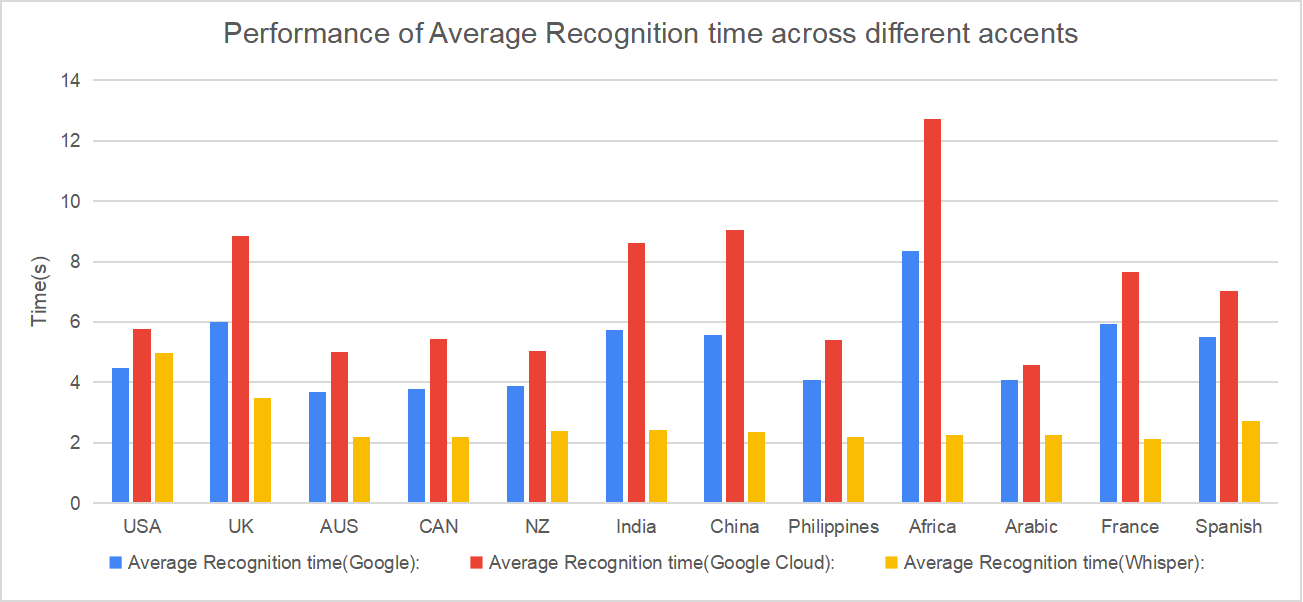}
    \caption{Performance of Average Recognition Time across different accents.
}
    \label{fig:5}
\end{figure}

When examining the performance (Fig.~\ref{fig:5}) in terms of the average recognition time for these three speech-to-text APIs, Whisper consistently outperformed the others by consistently achieving the shortest processing time. This indicates that Whisper is the most efficient in converting speech into text.

Among native English-speaking countries, it was apparent that the UK had the lengthiest recognition time, whereas Australia had the shortest. Among non-native English-speaking countries, Africa demonstrated the lengthiest recognition time, while Arabic exhibited the shortest.

\begin{figure}[!ht]
    \centering
    \includegraphics[scale=0.48]{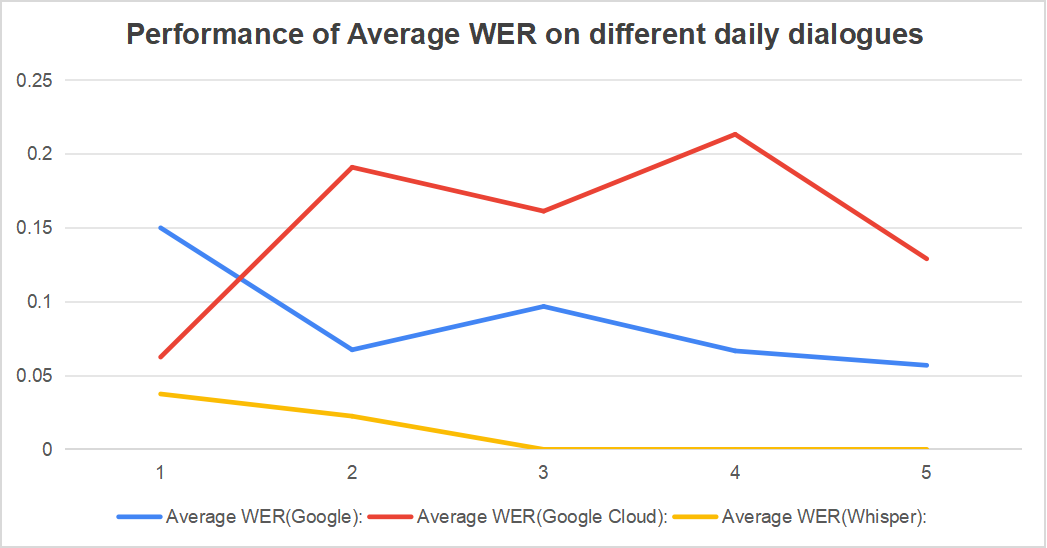}
    \caption{Performance of Average WER on different daily dialogues.
}
    \label{fig:6}
\end{figure}

Daily dialogue datasets were selected for our second test, comprising five conversations for which we recorded the audio. Then, this audio was processed through the three speech-to-text APIs and evaluated their performance in average word error rate (WER) and average recognition time. The results, as depicted in Figure~\ref{fig:6}, confirm that Whisper consistently achieved the lowest WER, signifying the highest level of accuracy of 1.72\%.

\begin{figure}[!ht]
    \centering
    \includegraphics[scale=0.42]{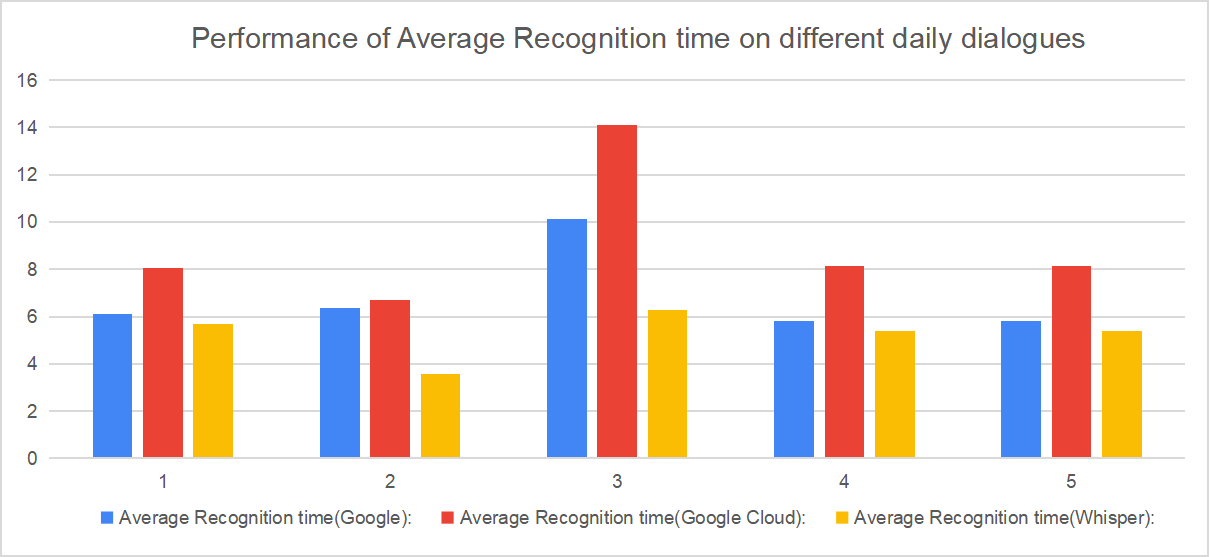}
    \caption{Performance of Average Recognition time on different daily dialogues.
}
    \label{fig:7}
\end{figure}

Lastly, in terms of Average Recognition Time(Figure~\ref{fig:7}), Whisper demonstrated the highest efficiency by consistently maintaining the shortest processing time at a number of 5.26 seconds.

Overall, Whisper has consistently outperformed the other two speech recognition APIs, thus affirming the suitability of our proposed method.

\subsection{Experiment Design}

Human trials are crucial for this research to ensure the ethical and practical implications of integrating ChatGPT with the Pepper Robot. The experiment involved students at The University of Auckland, with each participant engaging in an open-ended conversation with the integrated ChatGPT Pepper Robot. Each session lasted between 15 to 20 minutes. To recruit participants, our research was promoted by distributing flyers on designated poster boards within each department on the city campus. 

There were no specific criteria for potential participants, except that they should be able to communicate in English and be at least 18 years old. Participants were chosen on a first-come, first-served basis.

Obtaining informed consent is an essential ethical component of research involving human participants. It ensures that individuals fully comprehend the study's nature, objectives, potential advantages, risks, and rights as research participants. Consequently, all participants were requested to read through a "Participate Information Sheet" form and sign the "Consent form." This measure was taken to safeguard their privacy and confidentiality and to guarantee that the study adhered to ethical standards. During this process, researchers supplied participants with comprehensive information and addressed any inquiries they had, thus ensuring that participants possessed a thorough understanding of the study and were able to make an informed decision regarding their participation. Additionally, participants would have the option to withdraw from the study at any point. We inquired if participants were interested in receiving a copy of the study's findings to ensure transparency and offer participants the option to stay informed about the research outcomes.

Before the experiment commences, one of the researchers provided a brief overview of the integrated system's capabilities and features, along with recommended guidelines on how to initiate the conversation with the robot. Additionally, participants were furnished with microphones to enhance the clarity and accuracy of their recorded speech for improved speech recognition.

\begin{figure}[!ht]
    \centering
    \includegraphics[scale=0.4]{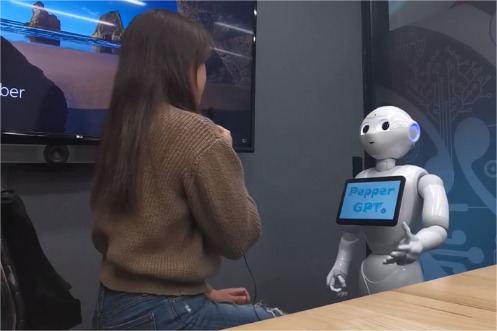}
    \caption{Interaction with the Pepper robot.}
    \label{fig: experiment}
\end{figure}

During the experiment (Fig.~\ref{fig: experiment}), participants were encouraged to engage in an open-ended conversation with the Pepper-GPT robot positioned in front of them. The system would transcribe the speech into text. For any potential technical support the participant needs during the experiment, one student researcher would be in the experiment room, seated discreetly in the corner. The interaction time with the robot was controlled between 5 and 10 minutes, depending on the participants' reactions.

Once the participant completed their interaction, they were asked to complete two digital questionnaires. The first questionnaire served as a baseline and collected information on participants' age, gender, academic department, ethical considerations, and prior experiences with ChatGPT. The second questionnaire gathered feedback on their experience interacting with the robot. Once participants completed these questionnaires, they were rewarded with a \$10 gift card. Subsequently, the collected responses were compiled and analyzed to derive findings.

\subsection{Quantitative Results}
The study involved 15 participants, with a gender distribution of six males and nine females. A significant portion of the participants, representing 60\%, fell within the age group of 18 to 23, as detailed in table \ref{table1}. 

\begin{figure}[!ht]
    \centering
    \includegraphics[scale=0.47]{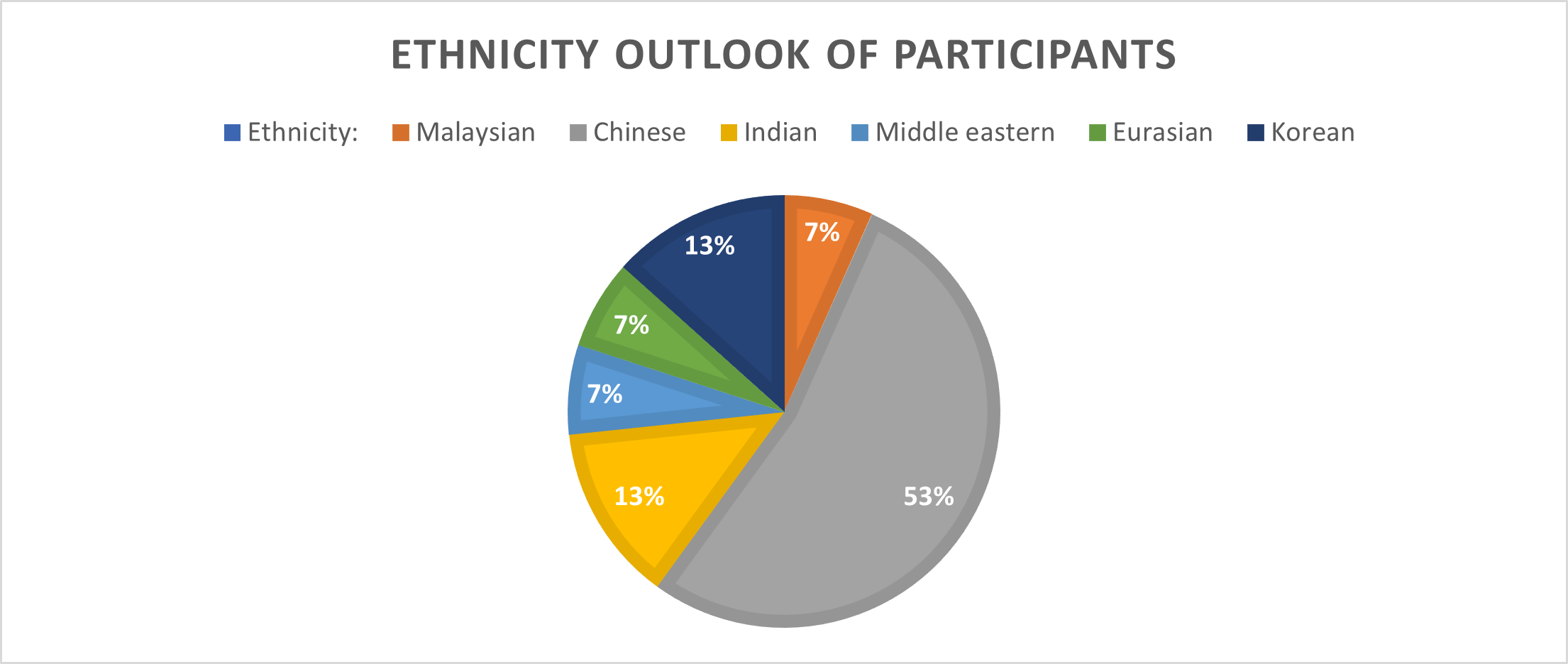}
    \caption{Ethnic group distribution among participants.
}
    \label{fig:8}
\end{figure}

Furthermore, the majority of participants, accounting for 53\%, were of Chinese origin, as depicted in Figure~\ref{fig:8}. 

\begin{figure}[!htb]
    \centering
    \includegraphics[scale=0.55]{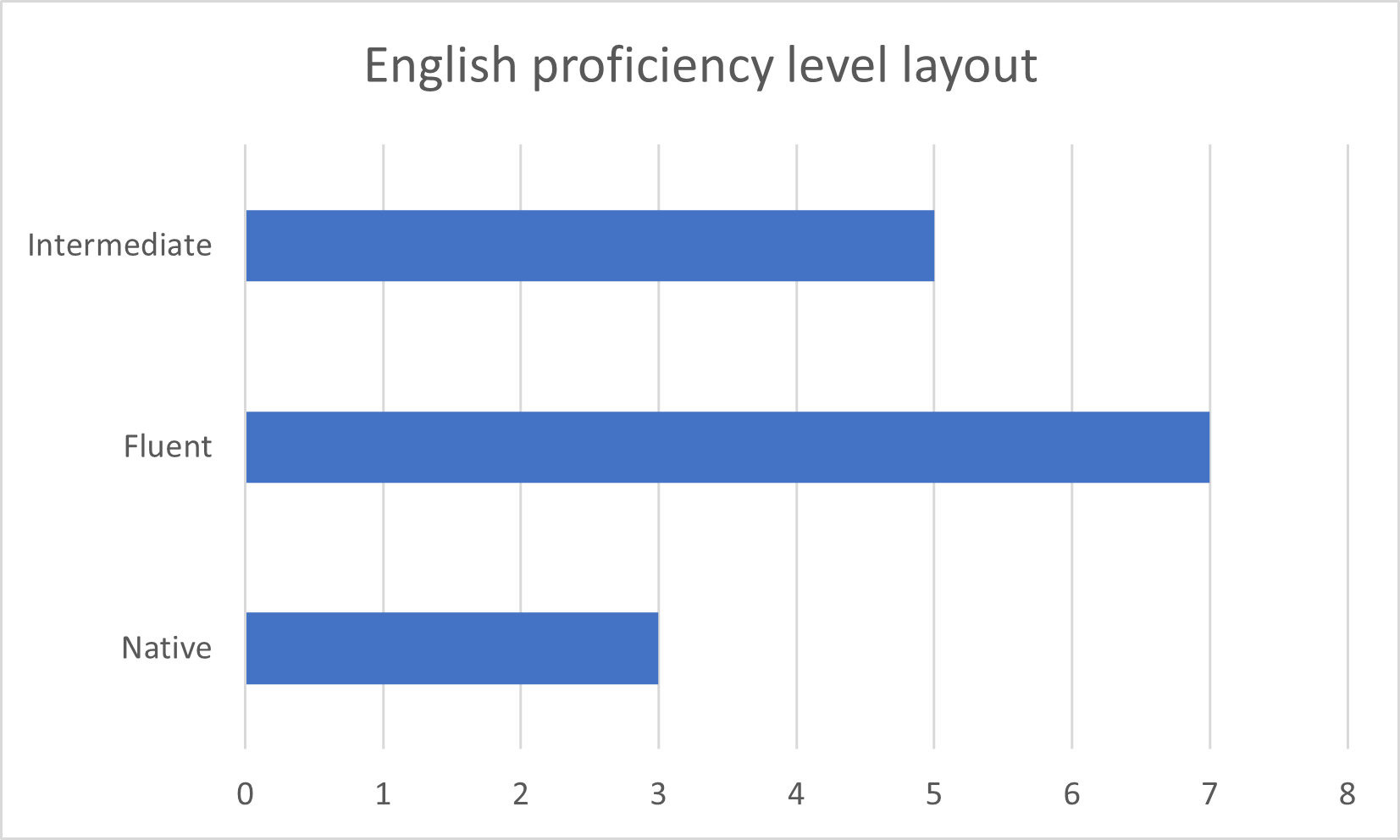}
    \caption{English proficiency level of participants.
}
    \label{fig:9}
\end{figure}

It is worth noting that more than half of the participants demonstrated a high level of English proficiency, indicating their capability to engage in smooth conversations with the robot, as illustrated in Figure~\ref{fig:9}. Additionally, a substantial number of participants were already familiar with ChatGPT and used it regularly.

\begin{figure}[!ht]
    \centering
    \includegraphics[scale=0.59]{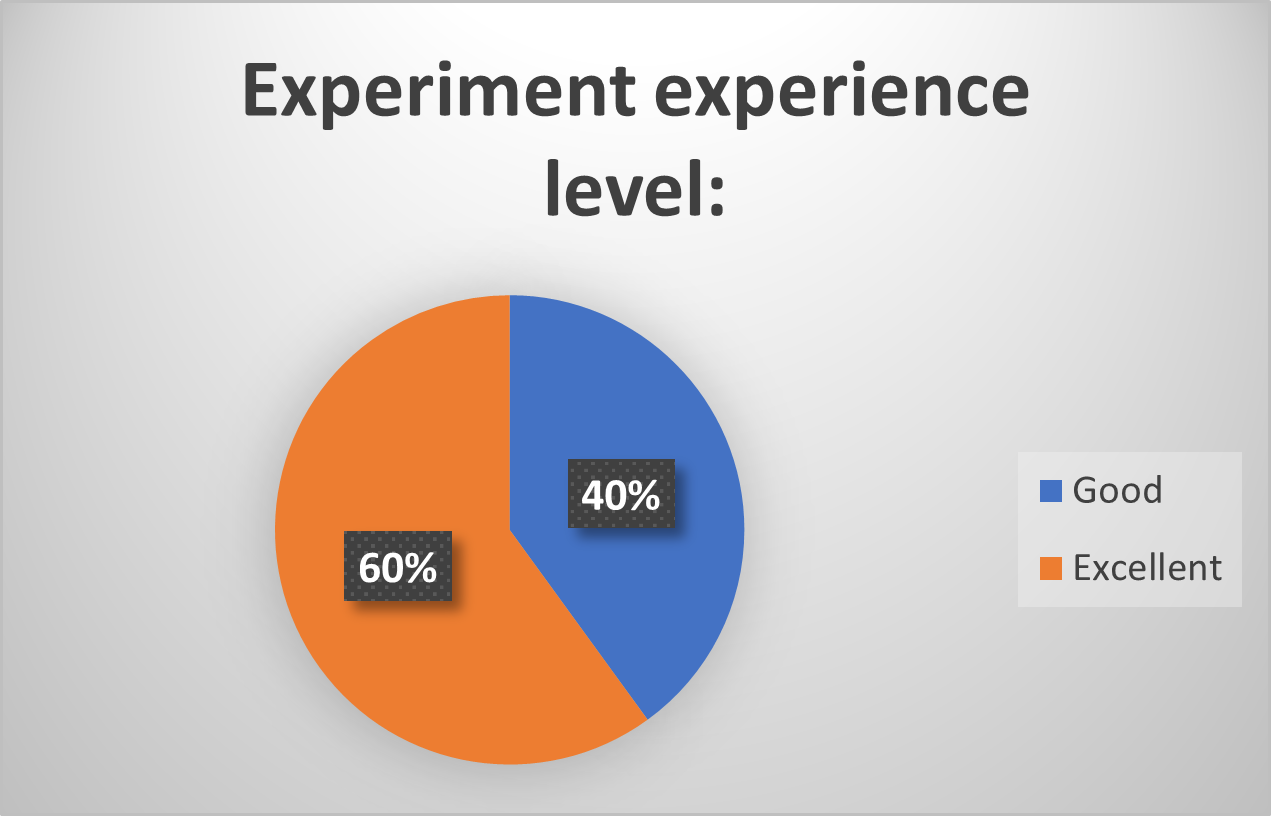}
    \caption{Experiment experience level of participants.}
    \label{fig:11}
\end{figure}

\begin{figure}[!ht]
    \centering
    \includegraphics[scale=0.56]{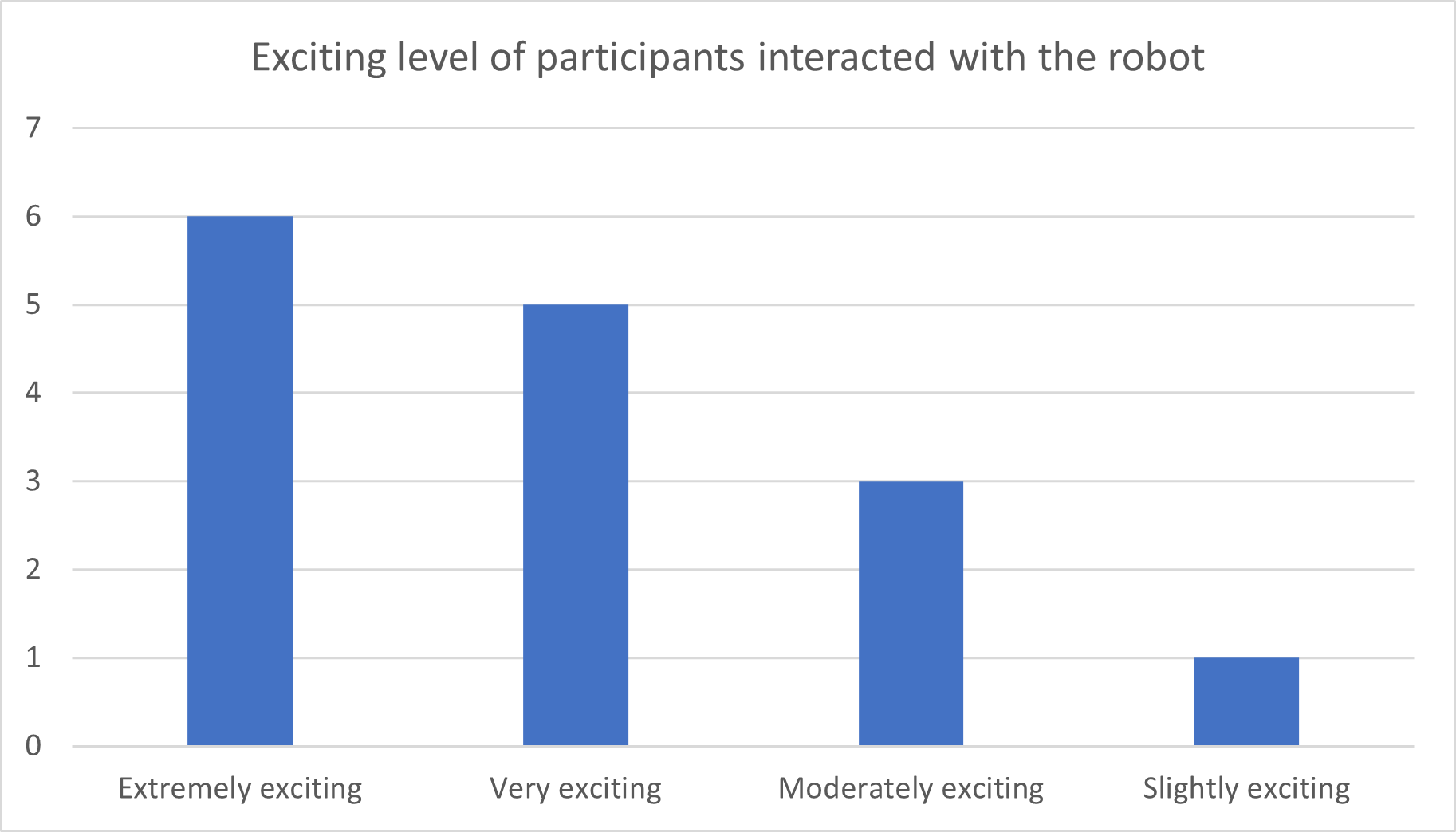}
    \caption{Excitement level of participants interacted with the robot.}
    \label{fig:12}
\end{figure}

\begin{figure}[!ht]
    \centering
    \includegraphics[scale=0.6]{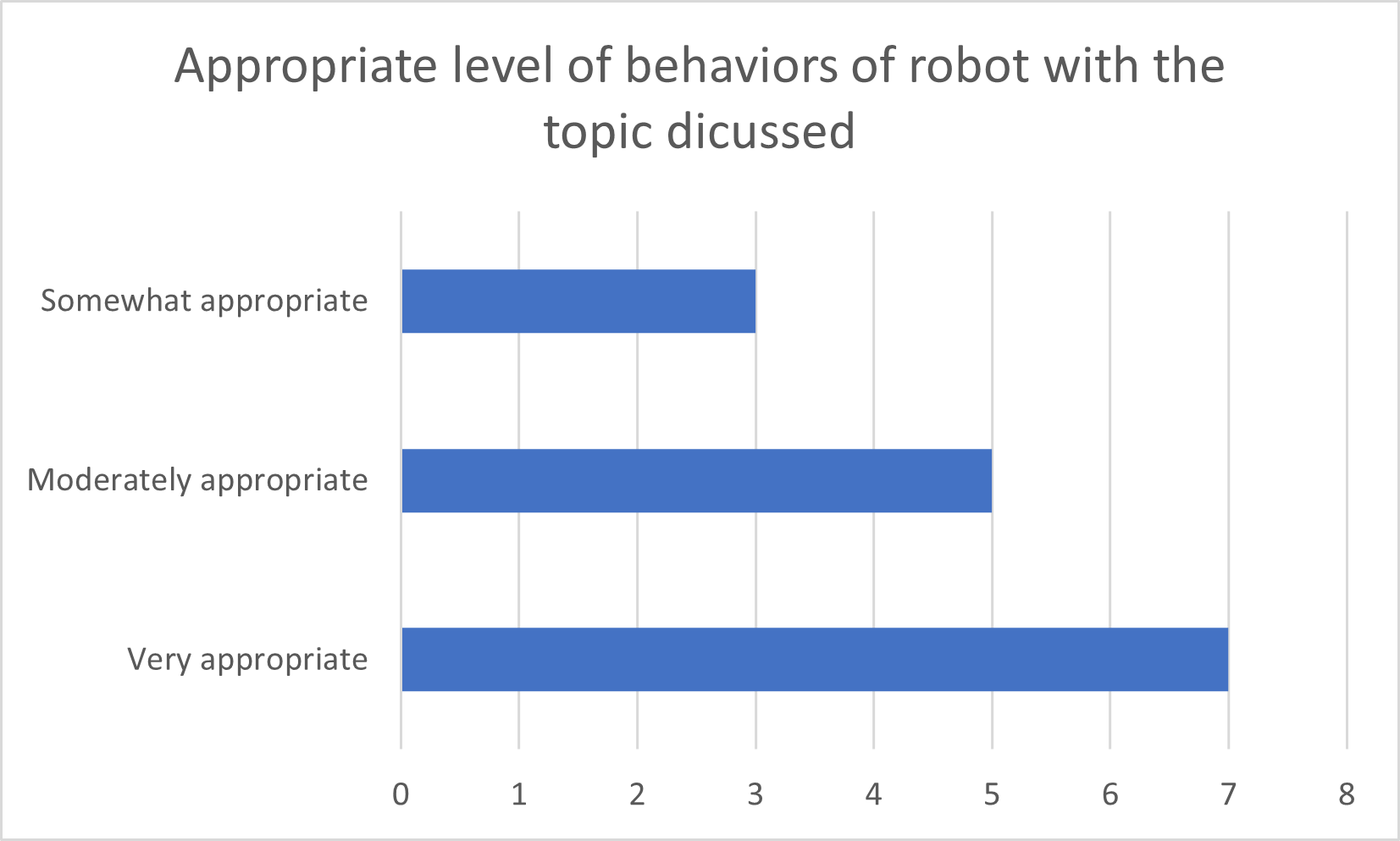}
    \caption{Appropriate level of behaviours of the robot with the topic discussed. }
    \label{fig:13}
\end{figure}

\begin{figure}[!ht]
    \centering
    \includegraphics[scale=0.65]{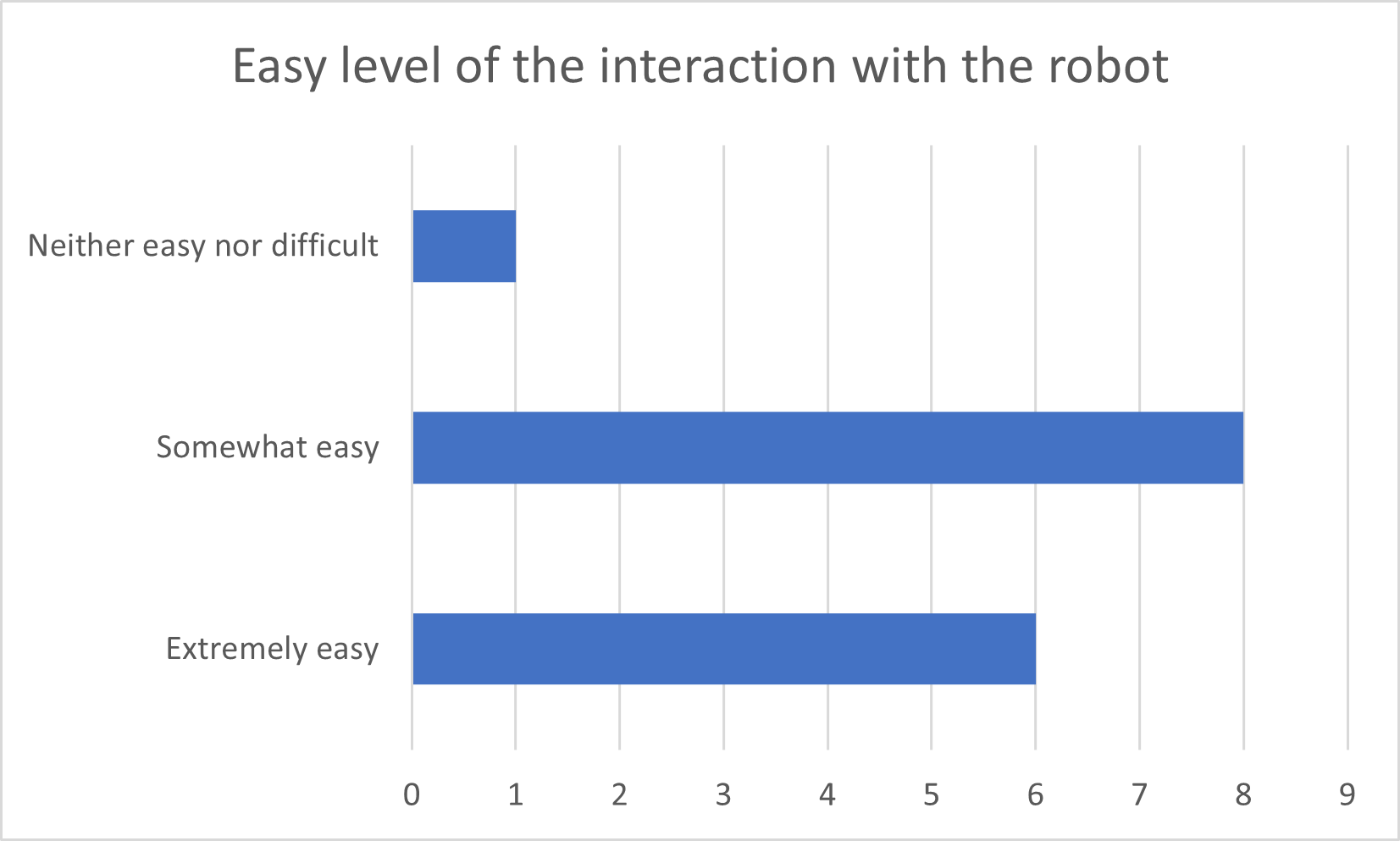}
    \caption{Easy level of the interaction with the robot.}
    \label{fig:14}
\end{figure}

Users’ perceptions of the system were explored, including their ratings of overall experience in this experiment(Figure~\ref{fig:11}), their ratings of the excitement level with the robot(Figure~\ref{fig:12}), the appropriateness of the robot’s gestures linked to the topics discussed(Figure~\ref{fig:13}) and the ease of interaction with the robot(Figure~\ref{fig:14}). 

\begin{table}[h]
        \caption{Age group of participants.}
        \label{table1}
        \begin{center}
            \begin{tabular}{|c||c|c|}
                \hline
                Participants age range & Proportion  \\
                \hline
                18 to 23 years & 60\%\\
                \hline
                24 to 29 years & 20\%\\
                \hline
               30 to 35 years & 20\%\\
                \hline
            \end{tabular}
        \end{center}
    \end{table}


The feedback from participants in the study reflected a generally positive experience with Pepper-GPT according to Figure~\ref{fig:11} as 60\% rated their overall experience as excellent. Over 70\% of the participants showed a high level of excitement in engaging in the conversation with the robot(Figure~\ref{fig:12}). Participants were pleased with the robot's ability to understand their questions and provide relevant answers since half of the results seemed they received a very appropriate response from the robot(Figure~\ref{fig:13}). The robot's attitude and physical behavior were well-received, enhancing the overall interaction. While some minor issues, such as connection problems and occasional gaps in knowledge, were noted, participants found the experience enjoyable and exciting.

In Figure~\ref{fig:14}, more than half of the participants did not think this integrated system was straightforward, which might be the development area for us.  
A few participants mentioned that they still preferred using text-based ChatGPT for convenience. Overall, the physical robot presence made ChatGPT interactions more realistic and engaging for most users.

\section{Discussion}

A notable correlation between Word Error Rate (WER) and processing time has been observed in speech recognition. To illustrate, consider the UK accent, which, despite exhibiting a higher WER, also incurred a lengthier processing time. This may be due to some UK accents featuring complex phonetic sounds and distinctions that are uncommon in other English accents. Conversely, the Australian accent, characterized by a lower WER, demonstrated a shorter processing time. While this pattern did not universally hold true across all cases, it unequivocally indicated a linear relationship between the two factors.

In our experiment, participants' overall experience appeared to be significantly influenced by their level of English proficiency. While Whisper's speech recognition generally performed well in accent testing, individuals with lower English proficiency encountered challenges during interactions with the robot. For example, they often rephrased their questions multiple times to ensure the robot accurately understood their intentions.

Moreover, our findings indicated that 30\% of the participants who had extensive prior experience using ChatGPT had higher expectations for the robot's performance compared to those who only occasionally used ChatGPT. This elevated expectation seemed to lead to some disappointment in the overall functionality of the system by the end of the experiment.

Furthermore, it is worth noting that participants who struggled with determining when to engage with the robot found the system less user-friendly. Additionally, the poor face tracking of the Pepper robot posed an additional challenge, as participants often had to repeatedly seek the robot's attention to establish eye contact.

In summary, a potential linear relationship exists between Word Error Rate and recognition time. English proficiency, user expectations, clarity of interaction timing, and face-tracking capabilities of the Pepper-GPT collectively influenced participants' experiences during the experiment. These factors highlight important areas for improvement in future iterations of the system to enhance overall user satisfaction and engagement.

\section{Future Work and Conclusion}
In conclusion, integrating Whisper ASR and GPT-3.5 APIs with the Pepper robot has significantly enhanced the user experience, bridging the gap between virtual AI and physical robots. The result from the ASR comparison proves that Whisper ASR got the best performance with an average Word Error Rate (WER) of 1.716\% and an average processing time of 2.639 seconds compared to the Google ASR and Google Cloud ASR, increasing the comprehension ability of the Pepper-GPT. The GPT module enables the robot to generate contextually relevant responses, detect user action commands, and execute corresponding instructions, making the interactions more diverse and attractive to users. In addition to verbal communication, robots are also good at displaying appropriate gestures according to the content of the dialogue to enrich the dialogue experience. 

The results of investigations from human participants strongly prove the potential for Pepper-GPT in the HRI field. Over 90\% of participants thought the system was user-friendly, and more than 50\% of them found that the gesture of the robot was appropriate. The feedback from their investigations demonstrates that participants were enjoying and interested in the Pepper-GPT and expected more future interactions with this system. 

While the Pepper-GPT has undoubtedly improved the user experience, several limitations have emerged during the experiments and require improvement to make the Pepper-GPT more natural, process efficient, and enjoyable during the interactions, further increasing the user experience:

\begin{itemize}
    
\item \textbf{Listening Hint:} Participants needed clarification about when to converse and wait for responses from the robot during the experiments. Therefore, providing more apparent cues to users about when the Pepper-GPT is listening to their speech can prevent users from confusion, lead to a smoother interaction, and enhance the user experience.
\item \textbf{Multilingual Ability:} The Pepper robot's language setup requirement restricts multilingual conversation. Future work should explore ways to enable the robot to seamlessly switch between different languages during interactions to adapt to diverse user preferences and requirements.
\item \textbf{More Actions Designed:} All the physical actions the Pepper-GPT can execute are predefined, and the number of these actions is limited. Increasing the number of executable physical actions for the robot will make the interaction more engaging and interactive.
\item \textbf{Facial Tracking Enhancement:} The performance of the Pepper robot's facial tracking API falls short of expectations as it has trouble capturing and following users’ faces accurately during conversations. Enhancing this feature would contribute to more natural and engaging interactions.
\end{itemize}

\bibliographystyle{named}
\bibliography{ref}

\end{document}